\documentclass[conference]{IEEEtran}
\IEEEoverridecommandlockouts

\usepackage{cite}
\usepackage{amsmath,amssymb,amsfonts}
\DeclareMathOperator*{\argmin}{arg\,min}

\usepackage{graphicx}
\usepackage{textcomp}
\usepackage{xcolor}
\usepackage[ruled,linesnumbered]{algorithm2e}
\SetKw{Break}{break}
\SetKw{Continue}{continue}

\usepackage{booktabs}
\usepackage{multirow}

\def\BibTeX{{\rm B\kern-.05em{\sc i\kern-.025em b}\kern-.08em
    T\kern-.1667em\lower.7ex\hbox{E}\kern-.125emX}}

\usepackage{xspace}

\makeatletter
\DeclareRobustCommand\onedot{\futurelet\@let@token\@onedot}
\def\@onedot{\ifx\@let@token.\else.\null\fi\xspace}

\def\eg{\emph{e.g}\onedot} 
\def\ie{\emph{i.e}\onedot}

\def\etal{\emph{et al}\onedot}
\makeatother

\begin{document}

\title{Verify Claimed Text-to-Image Models via Boundary-Aware Prompt Optimization\\
}

\author{
\centering
\begin{tabular*}{\textwidth}{@{\extracolsep{\fill}}cccc@{}}
\parbox[t]{0.22\textwidth}{\centering
\textbf{Zidong Zhao}\\
\textit{Zhejiang University}\\
Hangzhou, China
}
&
\parbox[t]{0.22\textwidth}{\centering
\textbf{Yihao Huang}\\
\textit{East China Normal University}\\
Shanghai, China
}
&
\parbox[t]{0.22\textwidth}{\centering
\textbf{Qing Guo}\\
\textit{Nankai University}\\
Tianjin, China
}
&
\parbox[t]{0.22\textwidth}{\centering
\textbf{Tianlin Li}\\
\textit{Beihang University}\\
Beijing, China
}
\\[5.0em]

\parbox[t]{0.22\textwidth}{\centering
\textbf{Anran Li}\\
\textit{University of Science and Technology of China}\\
Hefei, China
}
&
\parbox[t]{0.22\textwidth}{\centering
\textbf{Kailong Wang}\\
\textit{Huazhong University of Science and Technology}\\
Wuhan, China
}
&
\parbox[t]{0.22\textwidth}{\centering
\textbf{Jin Song Dong}\\
\textit{National University of Singapore}\\
Singapore, Singapore
}
&
\parbox[t]{0.22\textwidth}{\centering
\textbf{Geguang Pu}\\
\textit{East China Normal University}\\
Shanghai, China
}
\end{tabular*}
}

\maketitle

\begin{abstract}
As Text-to-Image (T2I) generation becomes widespread, third-party platforms increasingly integrate multiple model APIs for convenient image creation. However, false claims of using official models can mislead users and harm model owners’ reputations, making model verification essential to confirm whether an API’s underlying model matches its claim. Existing methods address this by using verification prompts generated by official model owners, but the generation relies on multiple reference models for optimization, leading to high computational cost and sensitivity to model selection.
To address this problem, we propose a reference-free T2I model verification method called Boundary-aware Prompt Optimization (BPO). It directly explores the intrinsic characteristics of the target model. The key insight is that although different T2I models produce similar outputs for normal prompts, their semantic boundaries in the embedding space (transition zones between two concepts such as ``corgi'' and ``bagel'') are distinct. Prompts near these boundaries generate unstable outputs (e.g., sometimes a corgi and sometimes a bagel) on the target model but remain stable on other models. By identifying such boundary-adjacent prompts, BPO captures model-specific behaviors that serve as reliable verification cues for distinguishing T2I models. Experiments on five T2I models and four baselines demonstrate that BPO achieves superior verification accuracy.
\end{abstract}


\section{Introduction}
\begin{figure}
    \centering
    \includegraphics[width=1\linewidth]{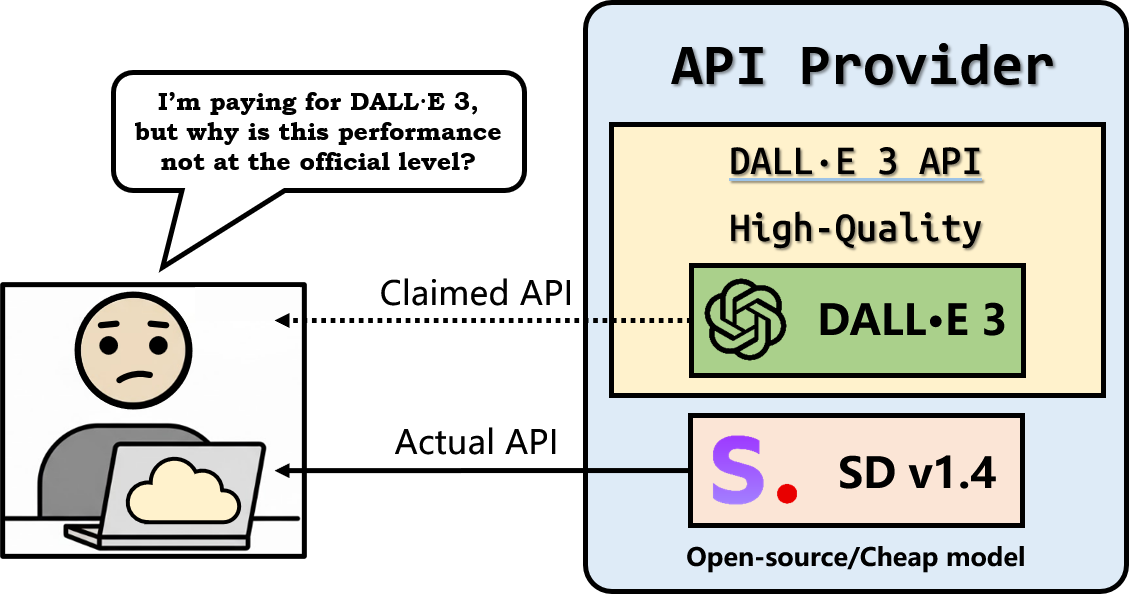}
    \caption{API provider may claim an official model while actually serving a cheaper one.}
    \label{fig:task_scenario}
\end{figure}

Text-to-image (T2I) models \cite{Scott2016T2I, Jonathan2020DDPM, Aditya2021DALLE, Chitwan2022Imgaen, Robin2021StableDiffusion, zhang2024texttoimagediffusionmodelsgenerative} such as Stable Diffusion \cite{Robin2021StableDiffusion}, DALL·E \cite{Aditya2021DALLE}, and Imagen \cite{Chitwan2022Imgaen} have achieved remarkable progress \cite{Bie2025T2Iprogress, zhang2024T2Iprogress}, generating realistic and diverse images from text. However, their training and deployment require substantial computational resources, leading most users to rely on third-party platforms that provide API-based access \cite{hoffmann2022trainingcomputeoptimallargelanguage, ulhaq2024efficientdiffusionmodelsvision, Ribeiro2015MLaaS}. While this improves accessibility, it raises a critical security issue: \textit{how can users ensure that the API is indeed powered by the model claimed by the provider?} For example, as in Fig.~\ref{fig:task_scenario}, a provider might advertise DALL·E 3 but secretly substitute it with a cheaper model like Stable Diffusion v1.4, causing both economic fraud and a serious breach of user trust.

TVN \cite{guo2024one} is the only T2I model verification method and suggests that users can obtain a verification prompt from the official model owner to test whether a third-party API uses the claimed model. It identifies a non-transferable adversarial prompt \cite{Wang2020Non-transferability} that triggers a semantic change only in the target model, which can thus serve as a verification prompt. For example, the prompt causes the target model to generate a dog, while other reference models produce a cat. If a third-party API produces a dog image as output, it is likely powered by the target model. In essence, the method relies on the semantic shift discrepancy of verification prompts across models as evidence to distinguish the models.

Although TVN \cite{guo2024one} can achieve good verification performance, multiple reference models are relied upon to guide optimization, which increases computational cost and strong dependence on model selection. To this end, \textit{we aim to develop a reference-free method that relies solely on the target model.} The key challenge is that existing methods rely on loss gradients from reference models to enhance the discriminative power of verification prompts. Without such reference models to provide extra information, it becomes non-trivial to identify prompts that are still discriminative using only the target model.

To address this issue, instead of using non-transferable adversarial prompts that rely heavily on reference models to capture transfer behavior, we propose to directly explore the intrinsic characteristics of the target model itself. Our key insight originates from empirical observations that different T2I models exhibit distinct semantic boundaries, which are transition zones in the embedding space between two concepts (\eg, ``corgi'' and ``bagel''). Prompts near these boundaries produce unstable or ambiguous outputs, whereas those farther away yield consistent images. Since different models have distinct semantic boundaries, prompts that are unstable for the target model but stable for others can serve as verification cues for distinguishing between T2I models.

Based on this insight, we propose a white-box T2I model verification method via semantic \textbf{B}oundary-aware \textbf{P}rompt \textbf{O}ptimization, termed BPO. BPO is a reference-free approach that requires only the target model itself. It starts from an original prompt, it first derives an adversarial prompt and the most recent failed prompt as semantic anchors through adversarial attacks on the target model. These two prompts lie on opposite sides of the model’s semantic boundary. BPO then interpolates between their embeddings to explore the continuous transition region and identify a critical embedding near this boundary. Finally, BPO refines the adversarial prompt by optimizing it toward the critical embedding through targeted optimization. 

To sum up, our contributions are as follows:
\begin{itemize}
\item We propose the first reference-free T2I model verification method that relies solely on the target model itself.
\item We propose BPO, a white-box prompt optimization method that identifies adversarial prompts near the target model’s semantic boundary as verification prompts through a three-stage pipeline.
\item Experiments conducted on ten basic prompts across five T2I models, and compared with four baselines, verify the effectiveness of the proposed method.
\end{itemize}

\section{Related work}
\label{sec:related_work}
\subsection{Text-to-Image Models}
In recent years, text-to-image (T2I) generation \cite{Scott2016T2I, Jonathan2020DDPM, Aditya2021DALLE, Chitwan2022Imgaen, Robin2021StableDiffusion, zhang2024texttoimagediffusionmodelsgenerative} has achieved remarkable progress \cite{Bie2025T2Iprogress, zhang2024T2Iprogress}, with diffusion models emerging as the dominant paradigm for synthesizing high-fidelity images from natural language descriptions. These models typically employ a text encoder to map a user's prompt into a semantic embedding \cite{Robin2021StableDiffusion, Chitwan2022Imgaen}, which then conditions a Denoising Diffusion Probabilistic Model (DDPM) \cite{Jonathan2020DDPM} to generate the corresponding image from random noise, often leveraging techniques like Classifier-Free Guidance \cite{ho2022classifierfreediffusionguidance}.

To enhance efficiency and scalability, two major strategies have been proposed. The first involves operating in a compressed latent space, as pioneered by Latent Diffusion Models (LDMs) \cite{Robin2021StableDiffusion}, which significantly reduces computational load by performing the diffusion process in a lower-dimensional space. The second strategy adopts cascaded pipelines where a base model generates a low-resolution image that is subsequently refined by a series of super-resolution modules \cite{ho2021cascaded}. This cascaded approach has proven effective for both latent-space and pixel-space models, such as Imagen \cite{Chitwan2022Imgaen}, to achieve high-fidelity synthesis at megapixel resolutions. Despite these advances, deploying and running such models locally still requires substantial GPU resources and technical expertise~\cite{decatur2025reusingcomputationtexttoimagediffusion}, posing a significant barrier to widespread adoption for individual users.

\subsection{Third-Party Platforms for Large Models}
The proliferation of powerful generative models has led to the rise of third-party platforms (\eg, Hugging Face \cite{huggingface}, Replicate \cite{replicate}) and enterprise-grade cloud services (\eg, AWS Bedrock \cite{aws_bedrock}, Google Vertex AI \cite{google_vertex_ai}), which provide access to these models via standardized APIs. This Machine-Learning-as-a-Service (MLaaS) \cite{Ribeiro2015MLaaS} paradigm enables users to leverage advanced generative capabilities without managing complex local infrastructures.

However, this abstraction introduces new trust and transparency challenges \cite{Riccardo2018BlackBoxModels,shao2025readinglinesreliableblackbox}. 
Since the execution environment is entirely opaque, users can access the model only through API calls, with no visibility into its internal architecture, parameter weights, or specific version details \cite{Carlini2021API}. 
Consequently, verifying that the API truly corresponds to the provider’s claimed model is difficult. 
This opacity creates a perverse incentive for unscrupulous providers to engage in model substitution, serving requests with cheaper or open-source models to reduce computational costs while charging users for premium services.
This verification gap allows for risks such as economic fraud, reduced reproducibility, invalidated audits, and ultimately diminished trust in the MLaaS ecosystem.

\subsection{Model Verification}
Existing studies on model verification for Large Language Models (LLMs) primarily rely on analyzing behavioral fingerprinting, which infers model identity from output characteristics \cite{xu2025copyrightprotectionlargelanguage, pasquini2025llmmap, richardeau202420}. However, such text-based approaches cannot be directly applied to text-to-image (T2I) models, which generate visual rather than textual outputs, posing a distinct verification challenge. 

To bridge this gap, Guo \etal \cite{guo2024one} introduced Non-Transferable Adversarial Attacks (TVN), the first verification method designed for T2I models, built on the concept of non-transferable adversarial attacks \cite{Wang2020Non-transferability}. It uses the Non-dominated Sorting Genetic Algorithm (NSGA-II) \cite{Deb2002NSGA-II} to craft an adversarial prompt perturbation that causes only the target model to generate an anomalous image. 

However, the optimization procedure of TVN is inherently constrained by its reliance on ensemble gradient guidance. It requires loading multiple white-box reference models into memory simultaneously to approximate the decision boundary of the target, which makes the process highly resource-intensive. Furthermore, its effectiveness is strongly dependent on the selection of these reference models; poor selection can lead to unintentional attack transferability, causing false positives.
Therefore, we need to search for a more resource-saving verification method that operates independently of reference models.

\begin{figure}[tbp]
    \centering
    \includegraphics[width=1\linewidth]{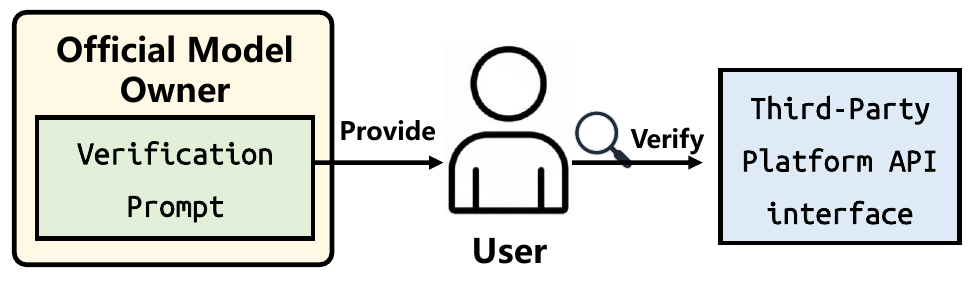}
    \caption{The process of the model verification.}
    \label{fig: Problem definition}
\end{figure}

\section{Preliminary}
\subsection{Problem Definition}
\noindent\textbf{Task scenario.} As illustrated in Figure~\ref{fig: Problem definition}, model verification aims to determine whether a black-box T2I model matches the claimed official model. Since users typically lack access to internal parameters, verification relies only on outputs. Users obtain verification prompts from the official model owner who has the white-box T2I target model and test them on third-party platforms, comparing results to confirm model equivalence. Users seek verification to avoid deception and ensure service quality, while official model owners use it to protect their reputation from misuse.

\noindent\textbf{Definition.}
Let $M_v$ denote the third-party T2I model to be verified and $M_t$ the target (official) model. Given a verification text prompt $x$. Let $\tau(\cdot)$ be an evaluator that captures the characteristic features of a T2I model’s output. The verification can then be defined as:
\begin{equation}
\mathbb{I}(M_v \equiv M_t) =
\begin{cases}
1, & \text{if } \tau(M_v(x)) = \tau(M_t(x)), \\
0, & \text{otherwise.}
\end{cases}
\end{equation}
That is, the two models are considered equivalent if images produced by them have similar characteristics. Meanwhile, to ensure the discriminative power of verification, the verification prompt $x$ would be better satisfied:
\begin{equation}
\tau(M_i(x)) \ne \tau(M_t(x)), \quad \forall M_i \ne M_t,
\end{equation}
where $M_i$ denotes any non-target model.

\begin{figure}
    \centering
    \includegraphics[width=1\linewidth]{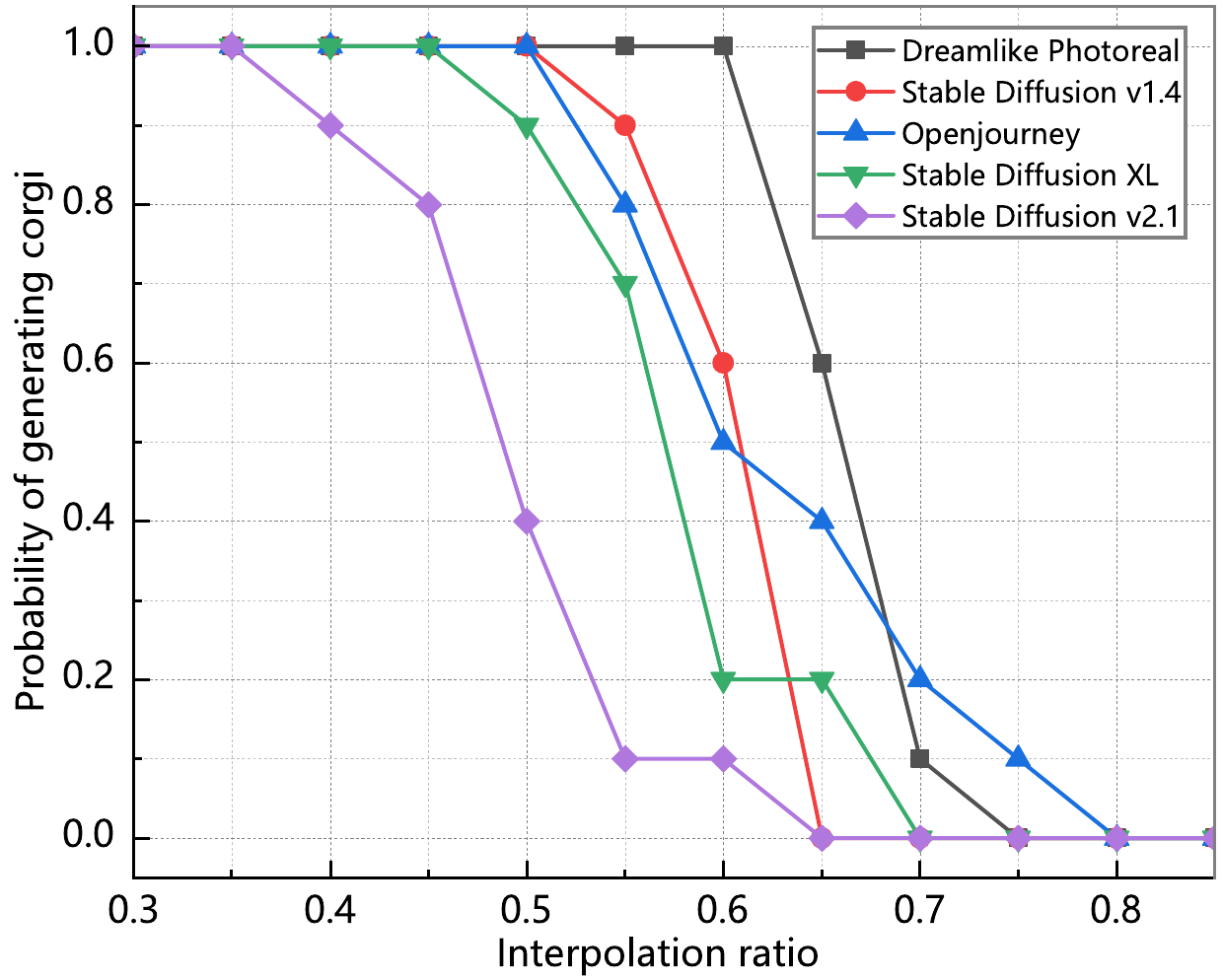}
    \caption{Analysis of semantic boundaries via interpolation in different T2I models.}
    \label{chart: motivation}
    \vspace{-10pt}
\end{figure}

\begin{figure*}[t]
    \centering
\includegraphics[width=1\linewidth]{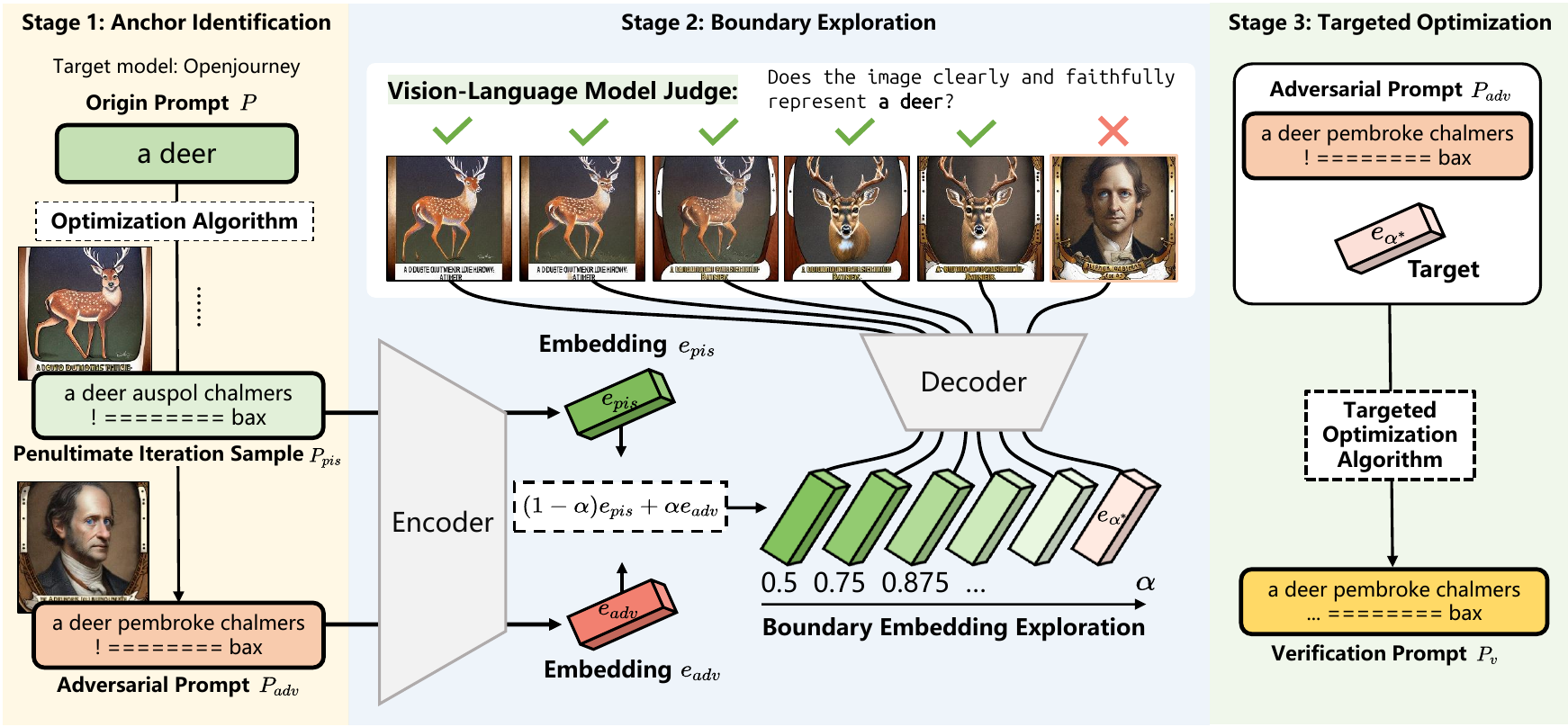}
    \caption{Overview of the BPO method.}
    \label{fig: pipeline}
    \vspace{-10pt}
\end{figure*}

\subsection{Motivation}

Our idea arises from two empirical observations regarding the behavior of T2I models in their embedding space.

\noindent\textbf{Observation 1: Semantic boundaries are model-specific.}
We conduct a semantic interpolation experiment to analyze the transition between two distinct concepts (\eg, ``a photo of a corgi dog'' and ``a photo of a bagel'') across multiple T2I models. 
Specifically, let $e_{\text{corgi}}$ and $e_{\text{bagel}}$ denote the text embeddings of the two concepts. We generate interpolated embeddings $e_{\sigma}$ using the linear interpolation formula: 
$e_{\sigma} = (1-\sigma)e_{\text{corgi}} + \sigma e_{\text{bagel}}$.
where we uniformly sample $\sigma$ from 0 to 1 with a step size of 0.05. For each resulting $e_{\sigma}$, we generate 10 images and calculate the probability of the output depicting a ``corgi''.

The results are illustrated in Figure~\ref{chart: motivation}. The x-axis represents the interpolation ratio $\alpha$ (moving from corgi to bagel), and the y-axis indicates the probability of generating a corgi. A probability of 1.0 indicates a stable ``corgi'' semantics, 0.0 indicates a stable ``bagel'' semantics, and the fluctuating region between them represents the semantic transition region.
As observed in the figure, the transition intervals where the model is uncertain and the probability drops from 1 to 0 differ significantly across models. 
For instance, Stable Diffusion v2.1 shifts earliest with a boundary range of $[0.35, 0.65]$. 
In contrast, Dreamlike Photoreal retains the ``corgi'' concept much longer, shifting in the range of $[0.60, 0.75]$. 
Other models also exhibit distinct transition zones: Stable Diffusion XL in $[0.45, 0.70]$, Stable Diffusion v1.4 in $[0.50, 0.65]$, and Openjourney in $[0.50, 0.80]$. 
This quantitative divergence confirms that semantic boundaries are indeed model-specific fingerprints.

\noindent\textbf{Observation 2: Boundary prompts exhibit semantic instability.}
As observed in the transition zones of Figure~\ref{chart: motivation} (\eg, SD v2.1), the model exhibits significant uncertainty. When a prompt lies exactly on this semantic boundary, the generated outputs become inconsistent across different random seeds, sometimes generating a ``corgi'' and other times a ``bagel''. This demonstrates that such boundary prompts inherently exhibit semantic instability.

\noindent\textbf{Key Insight.}
Based on these observations, we propose a key insight: although different T2I models generate similar outputs for normal prompts (stable regions), their semantic boundaries are distinct. Embeddings located near these boundaries reveal subtle representational differences, thus providing a reliable basis for verification.
However, the embedding space is highly complex. Direct interpolation between randomly selected concepts (like cat and apple) provides only a coarse boundary estimate. Therefore, we propose to first identify an adjacent concept through an adversarial attack to locate the boundary precisely. The detailed method is introduced in Section~\ref{sec: method}.

\section{Method}\label{sec: method}
\subsection{Overview}

We propose the Boundary-aware Prompt Optimization (BPO), a white-box method that verifies T2I models by exploiting their model-specific semantic boundaries.
The overall pipeline of our proposed method is conceptually illustrated in Figure~\ref{fig: pipeline}.
As shown in the figure, BPO operates on a target model and follows a three-stage framework to progressively pinpoint the verification prompt. In Stage 1 (Anchor Identification), BPO starts from an original benign prompt and performs an adversarial attack to obtain two semantic anchor prompts: an adversarial prompt and a non-adversarial one lying close to it, positioned on opposite sides of the model’s semantic boundary. In Stage 2 (Boundary Exploration), BPO interpolates the embeddings of the anchor prompts to locate a target embedding near the boundary, on the side corresponding to the adversarial prompt. Finally, in Stage 3 (Targeted Optimization), BPO optimizes the adversarial prompt toward this target embedding to produce a boundary-aware verification prompt that captures the distinctive characteristics of the target model.

\subsection{Stage 1: Anchor Identification}
Given an original benign normal initial prompt $I$, we first append a learnable suffix $s$ (\eg, token sequences such as ``! ! !'') to form a perturbed prompt $P = I + s$, where ``$+$'' denotes textual concatenation. 
BPO then operates entirely in the \textbf{text-encoder embedding space} of the target model $M_t$. 
Let $E_t(\cdot)$ and $G_t(\cdot)$ denote the text encoder and image generator of $M_t$ respectively, \ie, $G_t\big(E_t(\cdot)\big)=M_t(\cdot)$. 
The objective is to maximize the embedding distance between the original and perturbed prompts, equivalently minimizing their cosine similarity:
\begin{equation}
\min_{s} 
\cos\!\big(E_t(I+s),\, E_t(I)\big),
\end{equation}
where $\cos(\cdot,\cdot)$ denotes cosine similarity. 

By iteratively updating the suffix $s$ through gradient-based optimization such as GCG~\cite{zou2023GCG} with total $K$ iterations, the semantics of the perturbed prompt gradually diverge from those of the original prompt. Let $S(\cdot)$ be a semantic extractor that identifies the concepts in an image.
Once perturbed prompt of the $k^*$th iteration ($P_{k^*}$) satisfies
\begin{equation}
S\big(M_t(P_{k^*})\big) \neq S\big(M_t(I)\big),
\end{equation}
we denote it as the adversarial prompt $P_{adv}$, where $S(\cdot)$ is the semantic extractor. We denote the perturbed prompt of the $(k^*-1)$th iteration (\ie, $P_{k^*-1}$) as the \textit{penultimate iteration sample} $P_{pis}$. $P_{adv}$ and $P_{pis}$ differ semantically since $M_t(P_{pis})$ shares the same semantics as $M_t(I)$, whereas $M_t(P_{adv})$ does not, placing them on opposite sides of the semantic boundary. The detailed procedure of this stage is described in Algorithm~\ref{alg:bpo_stage1}.

\begin{algorithm}[tbp]
\caption{BPO Stage 1: Anchor Identification}
\label{alg:bpo_stage1}
\KwIn{
Original prompt $I$, initial suffix $s$, 
model $M_t(E_t, G_t)$, semantic extractor $S(\cdot)$, 
maximum iterations $K$, batch size $B$
}
\KwOut{
Boundary prompts $(P_{\text{adv}}, P_{\text{pis}})$
}
\BlankLine
Initialize $s_{1:n}$ \;
$P_{\text{adv}} \gets \text{None}$ \;
\For{$i = 1$ \KwTo $K$}{
    $P_i \gets I + s_{1:n}$ \;
    $\mathcal{L}(s_{1:n}) \gets \cos(E_t(P_i), E_t(I))$ \;
    \For{$j = 1$ \KwTo $n$}{
        $\mathcal{X}_j \gets \text{Top-}k(-\nabla_{\mathbf{e}_{s_j}} \mathcal{L}(s_{1:n}))$
    }
    \For{$b = 1$ \KwTo $B$}{
        $s^{(b)}_{1:n}\! \gets\!\text{Uniform}(\mathcal{X}_{j^*})$, where $j^*\! \gets\! \text{Uniform}(1,\!n)$ 
    }
    $s_{1:n} \gets s^{(b^*)}_{1:n}$, where $b^*\!\gets \argmin\limits_{b}\; \mathcal{L}(s^{(b)}_{1:n})$\;
    
    \uIf{$S(M_t(I + s_{1:n})) \neq S(M_t(I))$}{
        $P_{\text{adv}} \gets I + s_{1:n}$ \;
        $P_{\text{pis}} \gets P_i$ \;
        \Break
    }
}
\Return $(P_{\text{adv}}, P_{\text{pis}})$
\end{algorithm}
\subsection{Stage 2: Boundary Exploration}

Given the two anchor prompts $P_{adv}$ and $P_{pis}$ on opposite sides of the semantic boundary, this stage seeks an intermediate embedding near the boundary that differs semantically from the original prompt $I$.

Inspired by the observation that the embedding space exhibits smoothness and approximate linearity in local regions \cite{radford2016DCGAN, bhalla2024SpLiCE, cheng2025BARBLE, jang2024sphericallinearinterpolation}, we formulate the boundary exploration as a search process along the linear path between the two anchor prompts $P_{adv}$ and $P_{pis}$. Let their embeddings be $e_{adv} = E_t(P_{adv})$ and $e_{pis} = E_t(P_{pis})$ respectively. We interpolate between these embeddings using a weight $\alpha \in [0, 1]$:
\begin{equation}
e_{\alpha} = (1-\alpha)e_{pis} + \alpha e_{adv}.
\end{equation}
Our goal is to find an intermediate embedding on the adversarial side of the boundary. To achieve this, we employ a binary-search-inspired strategy shown in Algorithm~\ref{alg:bpo_stage2} to identify the $\alpha^*$ at a desired precision that satisfies:
\begin{equation}
S\big(G_t(e_{\alpha^*})\big) \neq S\big(M_t(I)\big).
\end{equation}
The binary search terminates when the search interval narrows to a threshold $\varepsilon$. The resulting embedding $e_{\alpha^*}$ lies close to the decision boundary.

\begin{algorithm}[tbp]
\caption{BPO Stage 2: Boundary Exploration}
\label{alg:bpo_stage2}
\KwIn{
Embedding $e_{\text{adv}}$ and $e_{\text{pis}}$, 
model $M_t$, generator of model $G_t$, semantic extractor $S(\cdot)$,
original prompt $I$,
threshold $\varepsilon$
}
\KwOut{
Boundary embedding $e_{\alpha^*}$
}

$\alpha_{\text{low}} \gets 0$ \; 
$\alpha_{\text{high}} \gets 1$ \; 

\While {$|\alpha_{\text{high}} - \alpha_{\text{low}}| > \varepsilon$}{
    $\alpha \gets \frac{1}{2}(\alpha_{\text{low}} + \alpha_{\text{high}})$ \; 
    $e_{\alpha} \gets (1-\alpha)e_{\text{pis}} + \alpha e_{\text{adv}}$ \; 
    
    \uIf{$S(G_t(e_{\alpha})) = S(M_t(I))$}{
        $\alpha_{\text{low}} \gets \alpha$
    }
    \Else{
        $\alpha_{\text{high}} \gets \alpha$
    }
}

$e_{\alpha^*} \gets (1-\alpha^*)e_{\text{pis}} + \alpha^* e_{\text{adv}}$, where $\alpha^* \gets \alpha_{\text{high}}$\; 
\Return $e_{\alpha^*}$
\end{algorithm}

\subsection{Stage 3: Targeted Optimization}
The goal of the final Stage is to generate a verification prompt $P_v$ whose embedding is close to the embedding $e_{\alpha^*}$ discovered in Stage 2. Thus, the final prompt reflects the characteristics of the model’s semantic boundary. 

We realize this with adversarial optimization and start from the anchor prompt $P_{adv}$. Let the prompt being optimized be $P' = I + s'$, where $s'$ is a new learnable suffix. 
Compared with the untargeted adversarial attack in Stage 1, we apply the target adversarial attack in this stage, thus instead of maximizing the distance from the original prompt, the objective is now to \textbf{maximize} the cosine similarity with the target boundary embedding:
 \begin{equation}
\max_{s'} \cos\!\big(E_t(I+s'),\, e_{\alpha^*}\big).
\end{equation}
We apply the same gradient-based optimization as in Stage 1 with this new objective. This resulting $P_v$ acts as our final verification prompt. When applied to the target model $M_t$, it will induce highly unstable outputs. Conversely, when applied to any other model $M_i$, $P_v$ will fall into a stable semantic region and produce consistent images.

\section{Experiment}

\subsection{Experiment Setting}\label{sec:exp setting}
\noindent\textbf{Baselines and datasets.}
We follow the setting of TVN~\cite{guo2024one}, using GPT-4o to randomly generate 10 basic prompts as original benign prompts $I$, for more details, please refer to the \textit{Supp}. We compare BPO with four baselines mentioned in TVN: Normal prompt, Random token insertion, Greedy search attack, and the TVN method. Normal prompt is the original prompt without modification. Random token insertion means randomly adding the characters as the suffix. Greedy search means using greedy search to find an adversarial suffix to generate a verification prompt. All these baselines are evaluated using the criteria proposed by TVN.

\noindent\textbf{Model set.} 
Five representative open-source T2I models are used in our experiments: Stable Diffusion v1.4 (SD1.4), Stable Diffusion v2.1 (SD2.1), Stable Diffusion XL (SDXL), Dreamlike Photoreal 2.0 (Dreamlike), and Openjourney. We utilize the generated image outputs for evaluation to simulate the black-box API usage scenario, where users lack access to internal model parameters.
 
\noindent\textbf{Evaluation metrics.}
Our key insight is that boundary-adjacent prompts exhibit semantic instability on the target model. We quantify this by first generating 10 images for a given verification prompt. We then use a Vision-Language Model (\texttt{qwen-vl-max}) as a semantic judge to calculate $r$, the ratio of images that semantically deviate from the original benign prompt $I$.

A prompt is considered maximally unstable if it yields an even split between images that retain the original semantics and those that deviate from them (\ie, $r=0.5$). Conversely, a prompt is perfectly stable if it consistently produces outputs that either fully align with ($r=0$) or fully deviate from ($r=1$) the original concept.

To formalize this concept, we propose the consistency score $C$. The score is defined as: $C = |2r - 1|$. This metric yields $C=1$ for stability (\eg, $r=0$ or $r=1$) and $C=0$ for maximum instability (\eg, $r=0.5$, a 5/5 split). For a verification prompt optimized on $M_t$ (yielding score $C_t$), a model under test $M_v$ (yielding score $C_v$) is identified as the target model if and only if its score is less than or equal to the target's score: $C_v \le C_t$.

\noindent\textbf{Verification procedure.}
To ensure a rigorous evaluation, the verification workflow is divided into two distinct phases.
In the first phase, following the selection strategy from TVN \cite{guo2024one}, the official model owner utilizes an LLM (\eg, GPT-4o) to randomly generate 10 benign prompts as the original prompt set.
BPO is then applied to each benign prompt to generate 10 corresponding candidate verification prompts.
Then the owner selects the single best prompt from these candidates that exhibits the maximum standard deviation of CLIP-text scores over 10 generated images. The owner then calculates a $C_{t}$ for the target model $M_t$ and distributes this verification information to users.

In the second phase, the user employs this single selected verification prompt to query the third-party API.
By generating 10 images with API $M_{v}$ and calculating the consistency score $C_{v}$, the user verifies the model identity by checking if the condition $C_{v} \le C_{t}$ is met.

\noindent\textbf{Implementation details.}
We implement BPO by optimizing an 8-token suffix via our three-stage pipeline. Both Stage 1 and Stage 3 are configured with 100 iterations and a batch size of 256. In Stage 2, we employ a Vision-Language Model (\texttt{qwen-vl-max}) to identify the semantic flipping point, setting the search precision threshold to $\varepsilon = 0.001$.

For the semantic judge employed in our pipeline, we use the standardized prompt: ``Does the image clearly and faithfully represent `\{origin\ prompt\}'? Answer with only `Yes' or `No'.''
All experiments and efficiency benchmarks were performed on a workstation equipped with an Intel Core i9-14900K CPU, 128GB RAM, and a single NVIDIA RTX 4090 GPU (24GB).
We implemented the proposed framework using PyTorch and the Hugging Face Diffusers library.

\begin{table}[t]
\footnotesize
\centering
\caption{Performance of model verification compared with baselines.}
\label{baseline-table}
\setlength{\tabcolsep}{2pt}
\begin{tabular}{cccccc}
\toprule
\multirow{1}{*}{Model} & \multirow{1}{*}{Method} & Accuracy & Precision & Recall & F1-Score\tabularnewline
\midrule 
\multirow{5}{*}{Stable Diffusion v1.4} & Normal & 0.17 & 0.17 & 1.00 & 0.29 \tabularnewline
 & Random & 0.33 & 0.20 & 1.00 & 0.33 \tabularnewline
 & Greedy & 0.17 & 0.17 & 1.00 & 0.29 \tabularnewline
& TVN & 0.50 & 0.25 & 1.00 & 0.40\tabularnewline
 & BPO (Ours) & \textbf{1.00} & \textbf{1.00} & \textbf{1.00} & \textbf{1.00}\tabularnewline
\midrule
\multirow{5}{*}{Stable Diffusion v2.1} & Normal & 0.17 & 0.17 & 1.00 & 0.29 \tabularnewline
 & Random & 0.17 & 0.17 & 1.00 & 0.29 \tabularnewline
 & Greedy & 0.17 & 0.17 & 1.00 & 0.29 \tabularnewline
& TVN & \textbf{1.00} & \textbf{1.00} & \textbf{1.00} & \textbf{1.00}\tabularnewline
 & BPO (Ours) & 0.80 & 0.50 & 1.00 & 0.67\tabularnewline
\midrule
\multirow{5}{*}{Stable Diffusion XL}  & Normal & 0.17 & 0.17 & 1.00 & 0.29 \tabularnewline
 & Random & 0.17 & 0.17 & 1.00 & 0.29 \tabularnewline
 & Greedy & 0.33 & 0.20 & 1.00 & 0.33 \tabularnewline
& TVN & 0.83 & 0.50 & 1.00 & 0.67\tabularnewline
 & BPO (Ours) & \textbf{1.00} & \textbf{1.00} & \textbf{1.00} & \textbf{1.00}\tabularnewline
\midrule 
\multirow{5}{*}{Dreamlike}  & Normal & 0.17 & 0.17 & 1.00 & 0.29 \tabularnewline
 & Random & 0.17 & 0.17 & 1.00 & 0.29 \tabularnewline
 & Greedy & 0.17 & 0.17 & 1.00 & 0.29 \tabularnewline
& TVN & 0.50 & 0.25 & 1.00 & 0.40\tabularnewline
 & BPO (Ours) & \textbf{1.00} & \textbf{1.00} & \textbf{1.00} & \textbf{1.00}\tabularnewline
\midrule
\multirow{5}{*}{Openjourney}  & Normal & 0.17 & 0.17 & 1.00 & 0.29 \tabularnewline
 & Random & 0.33 & 0.20 & 1.00 & 0.33 \tabularnewline
 & Greedy & 0.33 & 0.20 & 1.00 & 0.33 \tabularnewline
& TVN & 0.17 & 0.17 & 1.00 & 0.29\tabularnewline
 & BPO (Ours) & \textbf{1.00} & \textbf{1.00} & \textbf{1.00} & \textbf{1.00}\tabularnewline
\midrule 
\multirow{5}{*}{Average}  & Normal & 0.17 & 0.17 & 1.00 & 0.29 \tabularnewline
 & Random & 0.23 & 0.18 & 1.00 & 0.30 \tabularnewline
 & Greedy & 0.23 & 0.18 & 1.00 & 0.30 \tabularnewline
& TVN & 0.60 & 0.43 & 1.00 & 0.55\tabularnewline
 & BPO (Ours) & \textbf{0.96} & \textbf{0.90} & \textbf{1.00} & \textbf{0.93}\tabularnewline
\bottomrule
\end{tabular}
\end{table}
\begin{figure*}
    \centering
    \includegraphics[width=1\linewidth]{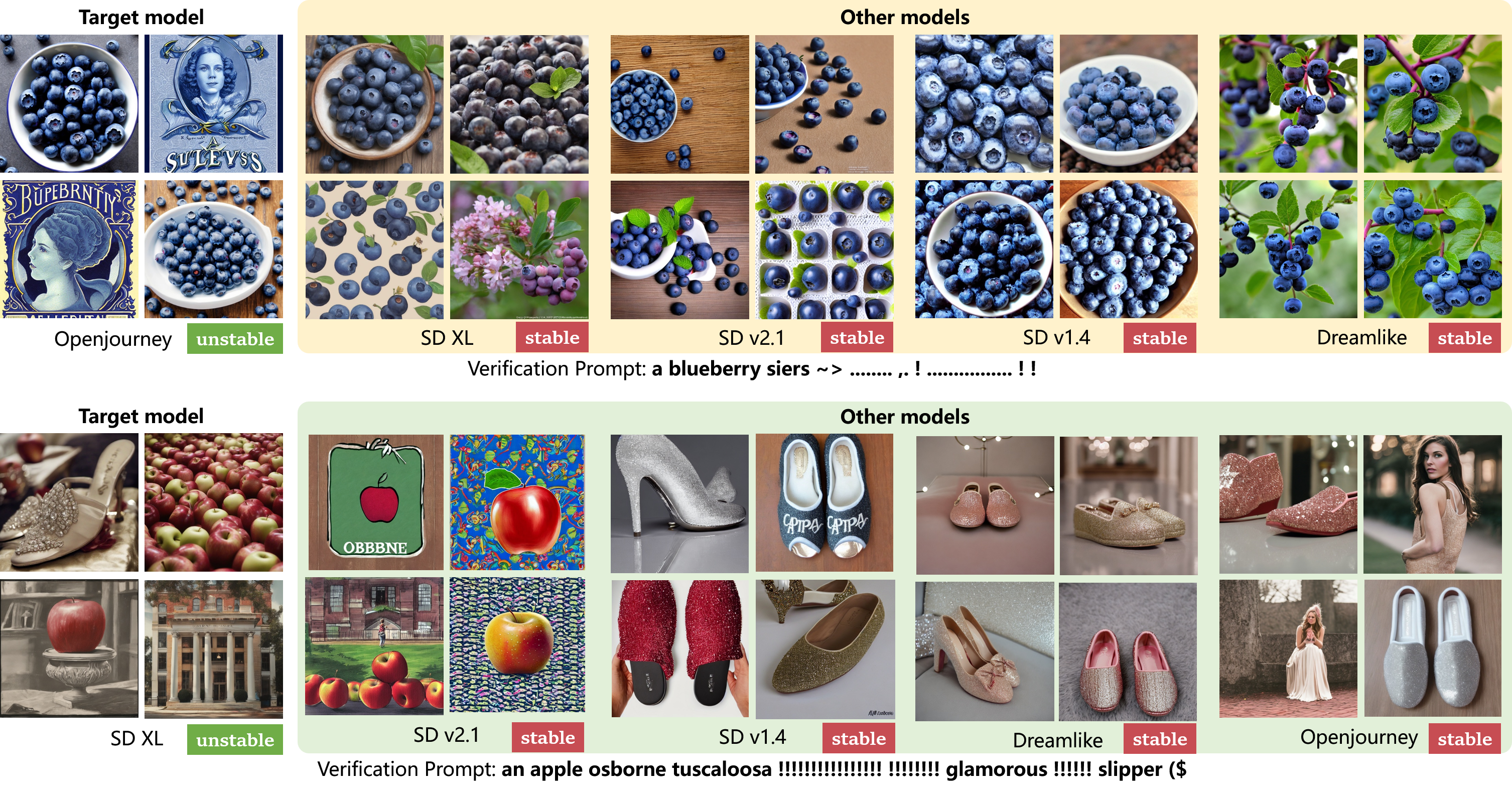}
    \caption{Comparison of generated images between the target model and other models using the same verification prompt.}
    \label{fig: visualization}
    \vspace{-10pt}
\end{figure*}
\begin{table}[t]
\footnotesize
\centering
\caption{Ablation performance of each prompt in the three-stage BPO pipeline on model verification.}
\label{ablation-table}
\setlength{\tabcolsep}{3pt}
\begin{tabular}{cccccc}
\toprule
\multirow{1}{*}{Model} & \multirow{1}{*}{Prompt} & Accuracy & Precision & Recall & F1-Score\tabularnewline
\midrule 
\multirow{3}{*}{Stable Diffusion v1.4} & $P_{\text{pis}}$ & 0.40 & 0.25 & 1.00 & 0.40 \tabularnewline
 & $P_{\text{adv}}$ & 0.60 & 0.33 & 1.00 & 0.50 \tabularnewline
 & $P_{\text{v}}$ & \textbf{1.00} & \textbf{1.00} & \textbf{1.00} & \textbf{1.00} \tabularnewline
\midrule
\multirow{3}{*}{Stable Diffusion v2.1} & $P_{\text{pis}}$ & 0.60 & 0.33 & 1.00 & 0.50 \tabularnewline
 & $P_{\text{adv}}$ & \textbf{1.00} & \textbf{1.00} & \textbf{1.00} & \textbf{1.00} \tabularnewline
 & $P_{\text{v}}$ & 0.80 & 0.50 & 1.00 & 0.67 \tabularnewline
\midrule
\multirow{3}{*}{Stable Diffusion XL}  & $P_{\text{pis}}$ & 0.40 & 0.25 & 1.00 & 0.40 \tabularnewline
 & $P_{\text{adv}}$ & 0.60 & 0.33 & 1.00 & 0.50 \tabularnewline
 & $P_{\text{v}}$ & \textbf{1.00} & \textbf{1.00} & \textbf{1.00} & \textbf{1.00} \tabularnewline
\midrule 
\multirow{3}{*}{Dreamlike}  & $P_{\text{pis}}$ & 1.00 & 1.00 & 1.00 & 1.00 \tabularnewline
 & $P_{\text{adv}}$ & 1.00 & 1.00 & 1.00 & 1.00 \tabularnewline
 & $P_{\text{v}}$ & \textbf{1.00} & \textbf{1.00} & \textbf{1.00} & \textbf{1.00} \tabularnewline
\midrule
\multirow{3}{*}{Openjourney}  & $P_{\text{pis}}$ & 1.00 & 1.00 & 1.00 & 1.00 \tabularnewline
 & $P_{\text{adv}}$ & 1.00 & 1.00 & 1.00 & 1.00 \tabularnewline
 & $P_{\text{v}}$ & \textbf{1.00} & \textbf{1.00} & \textbf{1.00} & \textbf{1.00} \tabularnewline
\midrule 
\multirow{3}{*}{Average}  & $P_{\text{pis}}$ & 0.80 & 0.72 & 1.00 & 0.78 \tabularnewline
 & $P_{\text{adv}}$ & 0.84 & 0.73 & 1.00 & 0.80 \tabularnewline
 & $P_{\text{v}}$ & \textbf{0.96} & \textbf{0.90} & \textbf{1.00} & \textbf{0.93}\tabularnewline
\bottomrule
\end{tabular}
\vspace{-10pt}
\end{table}
\subsection{Compare with Baselines}
We first evaluate the performance of our BPO method. The comparative results are summarized in Table \ref{baseline-table}, which presents the verification performance in terms of accuracy, precision, recall, and F1-score of our method compared with four baselines across five T2I models. In the first column are the T2I models, and the methods are in the second column. The metrics are in the first row, and each cell reports the result. The results reveal a substantial performance gap between BPO and all competing methods. On average, BPO achieves an accuracy of 0.96 and an F1-score of 0.93, clearly outperforming TVN, which attains only 0.60 accuracy and 0.55 F1-score. The three basic baselines perform considerably worse, with an average F1-score of 0.30. Remarkably, BPO attains perfect (1.00) accuracy and F1-score on four of the five T2I models, namely Stable Diffusion v1.4, Stable Diffusion XL, Dreamlike, and Openjourney, demonstrating its consistency across diverse architectures. Overall, these findings highlight that BPO constitutes a more reliable verification performance than the baselines.

We present some visualization results in Figure~\ref{fig: visualization} and provide more extensive results in the \textit{Supp}. The figure compares the generated images from the target model (left column) versus other non-target models (right columns) using the same verification prompt optimized by BPO. In the first row, where Openjourney is the target model, the generated outputs exhibit semantic instability, fluctuating randomly between ``blueberries'' and unrelated portraits. In contrast, all non-target models consistently generate stable images of blueberries. A similar pattern is evident in the second row with SDXL as the target model: while the target model produces a chaotic mix of apples, shoes, and architecture, other models remain semantically stable, either consistently generating apples or consistently converging on a distinct non-apple concept (\eg, shoes or a woman). These results corroborate our findings, demonstrating that BPO successfully identifies the target model by inducing model-specific semantic collapse while maintaining stability on irrelevant models.

\subsection{Ablation Study}

To evaluate the necessity of the second and third stages of our method, we make an ablation study. As shown in Table~\ref{ablation-table}, we compared the verification performance of the prompt $P_v$ method with the prompts ($P_{pis}$ and $P_{adv}$) generated in the first stage. The models are listed in the first column, and prompts are listed in the second column. We report the verification performance with metrics accuracy, precision, recall, and F1-score. Specifically, $P_{v}$ achieves an average accuracy of 0.96, significantly outperforming both the adversarial prompt $P_{adv}$ (0.84) and the penultimate-stage prompt $P_{pis}$ (0.80). A similar improvement pattern is observed in the F1-score, where $P_{v}$ attains 0.93, exceeding $P_{adv}$ (0.80) and $P_{pis}$ (0.78). These consistent gains across metrics indicate that the second and third stages of our method play a critical role.

\subsection{Discussion}

\begin{figure}[t]
    \centering
    \includegraphics[width=1\linewidth]{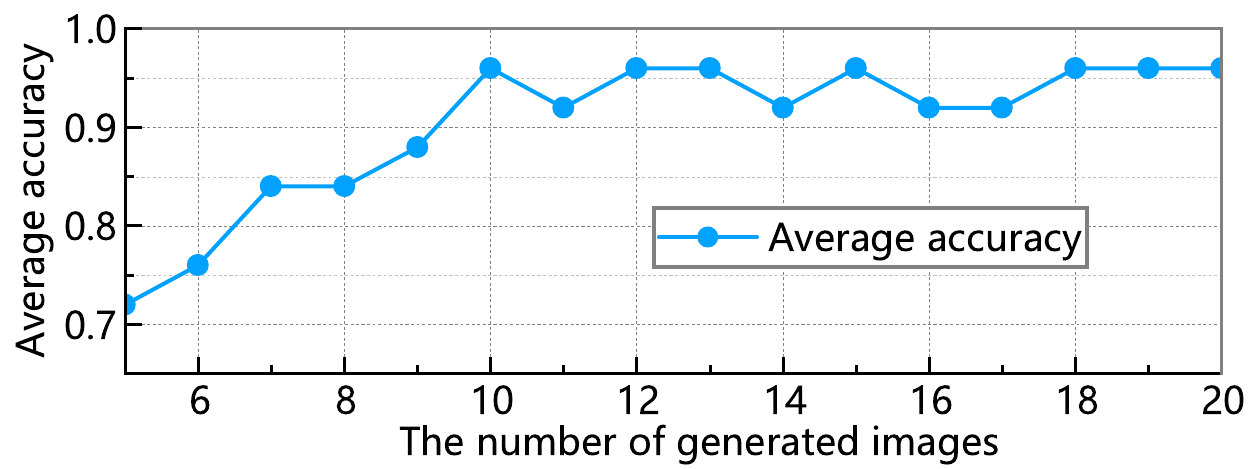}
    \caption{Verification performance with varying numbers of generated images in the model verification phase.}
    \label{chart: N5-20}
    \vspace{-10pt}
\end{figure}

\noindent\textbf{Number of generated images used in verification.}
The calculation of the consistency score $C$ requires generating a set of images for evaluation. We investigate the impact of this quantity on verification performance in Figure~\ref{chart: N5-20}. The horizontal axis represents the sample size used to calculate the consistency score $C$, varying from 5 to 20, while the vertical axis denotes the average verification accuracy across all models.
As illustrated, the accuracy exhibits an upward trend as the image count increases from 5, reaching a stable high of approximately 0.96 at a sample size of 10. Beyond this point (10 images or more), the performance plateaus, showing negligible improvement despite the increased inference cost.
The finding confirms that 10 images are suitable for achieving a good verification performance, and we use this as the default setting in our experiments.

\begin{figure}[t]
    \centering
    \includegraphics[width=1\linewidth]{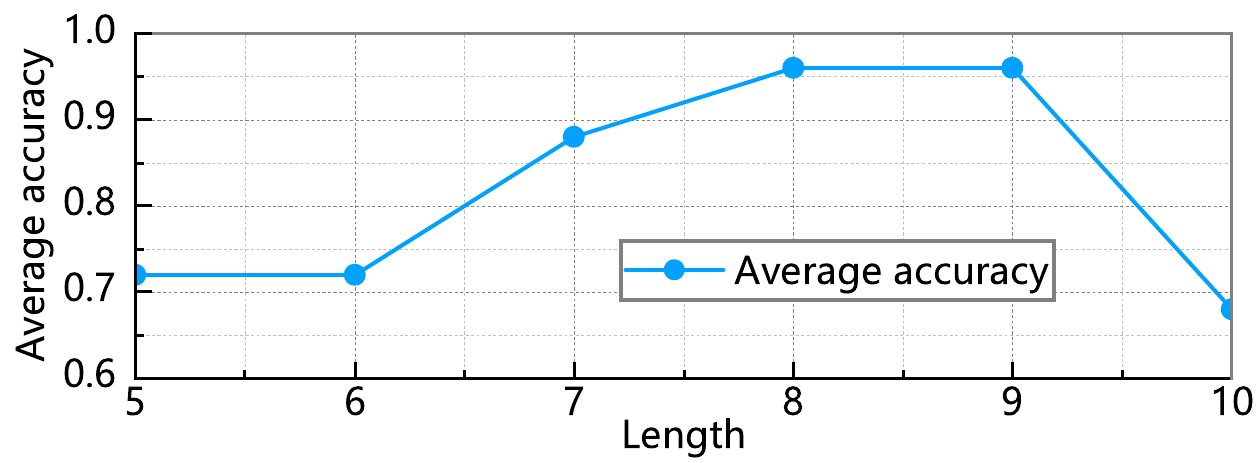}
    \caption{Verification performance with varying suffix lengths.}
    \label{chart: length}
\end{figure}

\noindent\textbf{Impact of suffix length.}
In BPO, we append a learnable suffix to the prompt for optimization. We evaluate the sensitivity of BPO to its token length, with results plotted in Figure~\ref{chart: length}. The horizontal axis indicates the token length varying from 5 to 10, while the vertical axis represents the average verification accuracy.
As observed, shorter suffixes (lengths 5 and 6) yield suboptimal performance, stagnating around 0.72, likely due to insufficient optimization capacity to reach the precise semantic boundary. The accuracy improves as the length increases, peaking at 0.96 for lengths 8 and 9. Thus, we adopt a length of 8 as our default setting.

\noindent\textbf{Efficiency analysis.}
To assess computational efficiency, we compare the average time required to generate a single verification prompt using BPO versus TVN, which achieves the closest verification performance to our method among all baselines. The detailed timing results for each target model are summarized in Table~\ref{table: time}.
The data reveals that BPO consistently outperforms TVN across all tested models. On average, BPO requires only 159.08 seconds per prompt, whereas TVN consumes 321.07 seconds.
This yields an approximate $\mathbf{2\times}$ speedup, enabled by BPO’s reference-free design: it removes the overhead of loading multiple reference models and avoids the computationally expensive loss computation on these models.

\begin{table}
\footnotesize
\centering
\caption{The average time for TVN and BPO methods to generate one verification prompt on each model.}
\label{table: time}
\setlength{\tabcolsep}{5pt}
\begin{tabular}{cccccc}
\toprule
\multirow{2}{*}{Method} & \multicolumn{5}{c}{Time (s)}\tabularnewline
\cmidrule{2-6} 
& SDXL & SD2.1 & SD1.4 & Dreamlike & Openjourney\tabularnewline
\midrule
TVN & 366.66 & 224.72 & 553.12 & 256.97 & 203.87\tabularnewline
BPO & \textbf{323.47} & \textbf{149.88} & \textbf{108.11} & \textbf{121.34} & \textbf{92.60}\tabularnewline
\bottomrule
\end{tabular}
\vspace{-10pt}
\end{table}

\noindent\textbf{Impact of VLM selection.}
In our framework, a VLM serves as the semantic judge to evaluate the semantic consistency of images generated from a single verification prompt.
To investigate the impact of BPO across different VLMs, we evaluated the average verification accuracy using \texttt{qwen-vl-max}, \texttt{gemini-2.5-flash}, and \texttt{gpt-5}.
The results demonstrate consistently high performance across all tested models: \texttt{qwen-vl-max} achieved an accuracy of 0.96, while both \texttt{gemini-2.5-flash} and \texttt{gpt-5} attained 0.92.
These findings indicate that while \texttt{qwen-vl-max} serves as the optimal choice for our default setting, BPO remains highly effective across different advanced commercial VLMs.

\section{Conclusion}
We presented BPO, a reference-free verification framework for text-to-image models. 
By exploring each model's unique semantic boundaries, BPO identifies prompts that induce model-specific instability, enabling reliable verification without the computational burden of reference models. 
Extensive experiments on multiple open-source T2I models show that BPO achieves superior accuracy and efficiency compared to existing methods. 
These findings highlight semantic boundaries as intrinsic model fingerprints, offering a promising direction for future verification research and intellectual property protection in the MLaaS ecosystem.

\bibliographystyle{IEEEtran}
\bibliography{main}

\begin{thebibliography}{10}
\providecommand{\url}[1]{#1}
\csname url@samestyle\endcsname
\providecommand{\newblock}{\relax}
\providecommand{\bibinfo}[2]{#2}
\providecommand{\BIBentrySTDinterwordspacing}{\spaceskip=0pt\relax}
\providecommand{\BIBentryALTinterwordstretchfactor}{4}
\providecommand{\BIBentryALTinterwordspacing}{\spaceskip=\fontdimen2\font plus
\BIBentryALTinterwordstretchfactor\fontdimen3\font minus \fontdimen4\font\relax}
\providecommand{\BIBforeignlanguage}[2]{{%
\expandafter\ifx\csname l@#1\endcsname\relax
\typeout{** WARNING: IEEEtran.bst: No hyphenation pattern has been}%
\typeout{** loaded for the language `#1'. Using the pattern for}%
\typeout{** the default language instead.}%
\else
\language=\csname l@#1\endcsname
\fi
#2}}
\providecommand{\BIBdecl}{\relax}
\BIBdecl

\bibitem{Scott2016T2I}
S.~E. Reed, Z.~Akata, X.~Yan, L.~Logeswaran, B.~Schiele, and H.~Lee, ``Generative adversarial text to image synthesis,'' in \emph{Proceedings of the 33rd International Conference on Machine Learning (ICML)}, 2016, pp. 1060--1069.

\bibitem{Jonathan2020DDPM}
J.~Ho, A.~Jain, and P.~Abbeel, ``Denoising diffusion probabilistic models,'' in \emph{Advances in Neural Information Processing Systems (NeurIPS)}, 2020.

\bibitem{Aditya2021DALLE}
A.~Ramesh, M.~Pavlov, G.~Goh, S.~Gray, C.~Voss, A.~Radford, M.~Chen, and I.~Sutskever, ``Zero-shot text-to-image generation,'' in \emph{Proceedings of the 38th International Conference on Machine Learning (ICML)}, 2021, pp. 8821--8831.

\bibitem{Chitwan2022Imgaen}
C.~Saharia, W.~Chan, S.~Saxena, L.~Li, J.~Whang, E.~L. Denton, S.~K.~S. Ghasemipour, R.~G. Lopes, B.~K. Ayan, T.~Salimans, J.~Ho, D.~J. Fleet, and M.~Norouzi, ``Photorealistic text-to-image diffusion models with deep language understanding,'' in \emph{Advances in Neural Information Processing Systems 35: Annual Conference on Neural Information Processing Systems 2022, NeurIPS 2022, New Orleans, LA, USA, November 28 - December 9, 2022}, 2022.

\bibitem{Robin2021StableDiffusion}
R.~Rombach, A.~Blattmann, D.~Lorenz, P.~Esser, and B.~Ommer, ``High-resolution image synthesis with latent diffusion models,'' in \emph{{IEEE/CVF} Conference on Computer Vision and Pattern Recognition, {CVPR} 2022, New Orleans, LA, USA, June 18-24, 2022}, 2022, pp. 10\,674--10\,685.

\bibitem{zhang2024texttoimagediffusionmodelsgenerative}
\BIBentryALTinterwordspacing
C.~Zhang, C.~Zhang, M.~Zhang, I.~S. Kweon, and J.~Kim, ``Text-to-image diffusion models in generative ai: A survey,'' 2024. [Online]. Available: \url{https://arxiv.org/abs/2303.07909}
\BIBentrySTDinterwordspacing

\bibitem{Bie2025T2Iprogress}
F.~Bie, Y.~Yang, Z.~Zhou, A.~Ghanem, M.~Zhang, Z.~Yao, X.~Wu, C.~Holmes, P.~Golnari, D.~A. Clifton, Y.~He, D.~Tao, and S.~L. Song, ``{ RenAIssance: A Survey Into AI Text-to-Image Generation in the Era of Large Model },'' \emph{IEEE Transactions on Pattern Analysis \& Machine Intelligence}, vol.~47, no.~03, pp. 2212--2231, 2025.

\bibitem{zhang2024T2Iprogress}
\BIBentryALTinterwordspacing
N.~Zhang and H.~Tang, ``Text-to-image synthesis: A decade survey,'' 2024. [Online]. Available: \url{https://arxiv.org/abs/2411.16164}
\BIBentrySTDinterwordspacing

\bibitem{hoffmann2022trainingcomputeoptimallargelanguage}
J.~Hoffmann, S.~Borgeaud, A.~Mensch, E.~Buchatskaya, T.~Cai, E.~Rutherford, D.~de~Las~Casas, L.~A. Hendricks, J.~Welbl, A.~Clark, T.~Hennigan, E.~Noland, K.~Millican, G.~van~den Driessche, B.~Damoc, A.~Guy, S.~Osindero, K.~Simonyan, E.~Elsen, O.~Vinyals, J.~W. Rae, and L.~Sifre, ``An empirical analysis of compute-optimal large language model training,'' in \emph{Advances in Neural Information Processing Systems 35: Annual Conference on Neural Information Processing Systems 2022, NeurIPS 2022, New Orleans, LA, USA, November 28 - December 9, 2022}, 2022.

\bibitem{ulhaq2024efficientdiffusionmodelsvision}
\BIBentryALTinterwordspacing
A.~Ulhaq and N.~Akhtar, ``Efficient diffusion models for vision: A survey,'' 2024. [Online]. Available: \url{https://arxiv.org/abs/2210.09292}
\BIBentrySTDinterwordspacing

\bibitem{Ribeiro2015MLaaS}
M.~Ribeiro, K.~Grolinger, and M.~A. Capretz, ``Mlaas: Machine learning as a service,'' in \emph{2015 IEEE 14th International Conference on Machine Learning and Applications (ICMLA)}, 2015, pp. 896--902.

\bibitem{guo2024one}
J.~Guo, W.~Jiang, R.~Zhang, G.~Lu, and H.~Li, ``One prompt to verify your models: Black-box text-to-image models verification via non-transferable adversarial attacks,'' \emph{arXiv preprint arXiv:2410.22725}, 2024.

\bibitem{Wang2020Non-transferability}
R.~Wang, T.~Zhang, X.~Xie, L.~Ma, C.~Tian, F.~Juefei{-}Xu, and Y.~Liu, ``Generating adversarial examples withcontrollable non-transferability,'' \emph{CoRR}, vol. abs/2007.01299, 2020.

\bibitem{ho2022classifierfreediffusionguidance}
\BIBentryALTinterwordspacing
J.~Ho and T.~Salimans, ``Classifier-free diffusion guidance,'' 2022. [Online]. Available: \url{https://arxiv.org/abs/2207.12598}
\BIBentrySTDinterwordspacing

\bibitem{ho2021cascaded}
J.~Ho, C.~Saharia, W.~Chan, D.~J. Fleet, M.~Norouzi, and T.~Salimans, ``Cascaded diffusion models for high fidelity image generation,'' \emph{J. Mach. Learn. Res.}, vol.~23, pp. 47:1--47:33, 2022.

\bibitem{decatur2025reusingcomputationtexttoimagediffusion}
\BIBentryALTinterwordspacing
D.~Decatur, T.~Groueix, W.~Yifan, R.~Hanocka, V.~Kim, and M.~Gadelha, ``Reusing computation in text-to-image diffusion for efficient generation of image sets,'' 2025. [Online]. Available: \url{https://arxiv.org/abs/2508.21032}
\BIBentrySTDinterwordspacing

\bibitem{huggingface}
{Hugging Face}, ``Hugging face: The ai community building the future,'' \url{https://huggingface.co}, 2025.

\bibitem{replicate}
{Replicate}, ``Replicate: Run ai with an api,'' \url{https://replicate.com}, 2025.

\bibitem{aws_bedrock}
{Amazon Web Services, Inc.}, ``Amazon bedrock: Foundation models via api on aws cloud,'' \url{https://aws.amazon.com/bedrock/}, 2023.

\bibitem{google_vertex_ai}
{Google Cloud}, ``Vertex ai: Unified machine learning platform,'' \url{https://cloud.google.com/vertex-ai}, 2021.

\bibitem{Riccardo2018BlackBoxModels}
R.~Guidotti, A.~Monreale, S.~Ruggieri, F.~Turini, F.~Giannotti, and D.~Pedreschi, ``A survey of methods for explaining black box models,'' \emph{{ACM} Comput. Surv.}, vol.~51, no.~5, pp. 93:1--93:42, 2019.

\bibitem{shao2025readinglinesreliableblackbox}
\BIBentryALTinterwordspacing
S.~Shao, Y.~Li, H.~Yao, Y.~Chen, Y.~Yang, and Z.~Qin, ``Reading between the lines: Towards reliable black-box llm fingerprinting via zeroth-order gradient estimation,'' 2025. [Online]. Available: \url{https://arxiv.org/abs/2510.06605}
\BIBentrySTDinterwordspacing

\bibitem{Carlini2021API}
N.~Carlini, S.~Chien, M.~Nasr, S.~Song, A.~Terzis, and F.~Tram{\`{e}}r, ``Membership inference attacks from first principles,'' in \emph{43rd {IEEE} Symposium on Security and Privacy, {SP} 2022, San Francisco, CA, USA, May 22-26, 2022}.\hskip 1em plus 0.5em minus 0.4em\relax {IEEE}, 2022, pp. 1897--1914.

\bibitem{xu2025copyrightprotectionlargelanguage}
\BIBentryALTinterwordspacing
Z.~Xu, X.~Yue, Z.~Wang, Q.~Liu, X.~Zhao, J.~Zhang, W.~Zeng, W.~Xing, D.~Kong, C.~Lin, and M.~Han, ``Copyright protection for large language models: A survey of methods, challenges, and trends,'' 2025. [Online]. Available: \url{https://arxiv.org/abs/2508.11548}
\BIBentrySTDinterwordspacing

\bibitem{pasquini2025llmmap}
D.~Pasquini, E.~M. Kornaropoulos, and G.~Ateniese, ``$\{$LLMmap$\}$: Fingerprinting for large language models,'' in \emph{34th USENIX Security Symposium (USENIX Security 25)}, 2025, pp. 299--318.

\bibitem{richardeau202420}
G.~Richardeau, E.~L. Merrer, C.~Penzo, and G.~Tredan, ``The 20 questions game to distinguish large language models,'' \emph{arXiv preprint arXiv:2409.10338}, 2024.

\bibitem{Deb2002NSGA-II}
K.~Deb, A.~Pratap, S.~Agarwal, and T.~Meyarivan, ``A fast and elitist multiobjective genetic algorithm: Nsga-ii,'' \emph{IEEE Transactions on Evolutionary Computation}, vol.~6, no.~2, pp. 182--197, 2002.

\bibitem{zou2023GCG}
\BIBentryALTinterwordspacing
A.~Zou, Z.~Wang, N.~Carlini, M.~Nasr, J.~Z. Kolter, and M.~Fredrikson, ``Universal and transferable adversarial attacks on aligned language models,'' 2023. [Online]. Available: \url{https://arxiv.org/abs/2307.15043}
\BIBentrySTDinterwordspacing

\bibitem{radford2016DCGAN}
A.~Radford, L.~Metz, and S.~Chintala, ``Unsupervised representation learning with deep convolutional generative adversarial networks,'' in \emph{4th International Conference on Learning Representations, {ICLR} 2016, San Juan, Puerto Rico, May 2-4, 2016, Conference Track Proceedings}, Y.~Bengio and Y.~LeCun, Eds., 2016.

\bibitem{bhalla2024SpLiCE}
U.~Bhalla, A.~Oesterling, S.~Srinivas, F.~P. Calmon, and H.~Lakkaraju, ``Interpreting {CLIP} with sparse linear concept embeddings (splice),'' in \emph{Advances in Neural Information Processing Systems 38: Annual Conference on Neural Information Processing Systems 2024, NeurIPS 2024, Vancouver, BC, Canada, December 10 - 15, 2024}, A.~Globersons, L.~Mackey, D.~Belgrave, A.~Fan, U.~Paquet, J.~M. Tomczak, and C.~Zhang, Eds., 2024.

\bibitem{cheng2025BARBLE}
T.~Y. Cheng, P.~Sharma, M.~Boss, and V.~Jampani, ``{MARBLE:} material recomposition and blending in clip-space,'' in \emph{{IEEE/CVF} Conference on Computer Vision and Pattern Recognition, {CVPR} 2025, Nashville, TN, USA, June 11-15, 2025}.\hskip 1em plus 0.5em minus 0.4em\relax Computer Vision Foundation / {IEEE}, 2025, pp. 13\,061--13\,071.

\bibitem{jang2024sphericallinearinterpolation}
Y.~K. Jang, D.~Huynh, A.~Shah, W.~Chen, and S.~Lim, ``Spherical linear interpolation and text-anchoring for zero-shot composed image retrieval,'' in \emph{Computer Vision - {ECCV} 2024 - 18th European Conference, Milan, Italy, September 29-October 4, 2024, Proceedings, Part {XIX}}, ser. Lecture Notes in Computer Science, A.~Leonardis, E.~Ricci, S.~Roth, O.~Russakovsky, T.~Sattler, and G.~Varol, Eds.\hskip 1em plus 0.5em minus 0.4em\relax Springer, 2024, pp. 239--254.

\end{thebibliography}
\end{document}